%% file: main.tex
\documentclass[conference]{IEEEtran}
\IEEEoverridecommandlockouts
\usepackage{cite}
\usepackage{amsmath,amssymb,amsfonts}
\usepackage{algorithmic}
\usepackage{graphicx}
\usepackage{textcomp}
\usepackage{xcolor}
\def\BibTeX{{\rm B\kern-.05em{\sc i\kern-.025em b}\kern-.08em
    T\kern-.1667em\lower.7ex\hbox{E}\kern-.125emX}}

\usepackage{booktabs}
\usepackage{tabularx}
\usepackage{bbold}
\newcolumntype{Y}{>{\centering\arraybackslash}X}
\usepackage{multirow}
\usepackage[spaces,hyphens]{xurl}
\usepackage{hyperref}

\makeatletter
\newcommand{\linebreakand}{%
  \end{@IEEEauthorhalign}
  \hfill\mbox{}\par
  \mbox{}\hfill\begin{@IEEEauthorhalign}
}
\makeatother
    
\begin{document}

\title{Heterogeneous Datasets for Federated Survival Analysis Simulation}

\author{
    \IEEEauthorblockN{Alberto Archetti}
    \IEEEauthorblockA{\textit{DEIB, Politecnico di Milano}\\
    Milan, Italy\\
    alberto.archetti@polito.it}
\and
    \IEEEauthorblockN{Eugenio Lomurno}
    \IEEEauthorblockA{\textit{DEIB, Politecnico di Milano}\\
    Milan, Italy\\
    eugenio.lomurno@polimi.it}
\and
    \IEEEauthorblockN{Francesco Lattari}
    \IEEEauthorblockA{\textit{DEIB, Politecnico di Milano}\\
    Milan, Italy\\
    francesco.lattari@polimi.it}
\linebreakand
    \IEEEauthorblockN{André Martin}
    \IEEEauthorblockA{\textit{Technische Universit{\"a}t Dresden}\\
    Dresden, Germany\\
    andre.martin@tu-dresden.de}
\and
    \IEEEauthorblockN{Matteo Matteucci}
    \IEEEauthorblockA{\textit{DEIB, Politecnico di Milano}\\
    Milan, Italy \\
    matteo.matteucci@polimi.it}
}

\maketitle

\input{src/0_abstract.tex}


\input{src/1_introduction.tex}
\input{src/2_related.tex}
\input{src/3_method.tex}
\input{src/4_experiments.tex}
\input{src/5_conclusion.tex}

\section*{Acknowledgment}
This project has been supported by AI-SPRINT: AI in Secure Privacy-pReserving computINg conTinuum (EU H2020 grant agreement No. 101016577).

\bibliographystyle{IEEEtran}
\bibliography{references}

\end{document}

%% file: src/0_abstract.tex
\begin{abstract}
Survival analysis studies time-modeling techniques for an event of interest occurring for a population. 
Survival analysis found widespread applications in healthcare, engineering, and social sciences.
However, the data needed to train survival models are often distributed, incomplete, censored, and confidential.
In this context, federated learning can be exploited to tremendously improve the quality of the models trained on distributed data while preserving user privacy.
However, federated survival analysis is still in its early development, and there is no common benchmarking dataset to test federated survival models.
This work provides a novel technique for constructing realistic heterogeneous datasets by starting from existing non-federated datasets in a reproducible way.
Specifically, we propose two dataset-splitting algorithms based on the Dirichlet distribution to assign each data sample to a carefully chosen client: quantity-skewed splitting and label-skewed splitting.
Furthermore, these algorithms allow for obtaining different levels of heterogeneity by changing a single hyperparameter.
Finally, numerical experiments provide a quantitative evaluation of the heterogeneity level using log-rank tests and a qualitative analysis of the generated splits.
The implementation of the proposed methods is publicly available in favor of reproducibility and to encourage common practices to simulate federated environments for survival analysis.
\end{abstract}

%% file: src/1_introduction.tex
\section{Introduction}

Survival analysis~\cite{klein2003survival,wang2019machine} is a subfield of statistics focused on modeling the occurrence time of an event of interest for a population.
In particular, its goal is to exploit statistical and machine learning techniques to provide a survival function, i.e., a function that estimates the event occurrence probability with respect to time for an individual.
Survival analysis has been successfully applied in many healthcare, engineering, and social science applications~\cite{emmert2019introduction}.
However, the data to train survival models are often distributed, incomplete, inaccurate, and confidential~\cite{andreux2020federated,rieke2020future}.
On top of that, survival data may include a considerable portion of censored observations, i.e., instances for which the event of interest has yet to occur.
In censored samples, the observed time is an underestimation of the actual occurrence time of the event.
As a result, data scarcity, censorship, and confidentiality can hinder the applicability of survival analysis when addressing real-world, large-scale problems.

In this context, Federated Learning (FL)~\cite{li2020federated,kairouz2021advances} holds tremendous potential to improve the effectiveness of survival analysis applications.
FL is a subfield of distributed machine learning that investigates techniques to train machine learning models while preserving user privacy.
In FL, data information never leaves the device in which it is produced, collected, and stored.
FL allows for training on large-scale data, improving the quality, fairness, and generalizability of the resulting models with respect to the non-distributed counterparts. 

Federated survival analysis studies the relationship between federated learning and survival analysis.
In particular, survival models present structural components that make their inclusion into existing federated learning algorithms non-trivial~\cite{andreux2020federated,wang2022survmaximin,lu2015webdisco,rahman2022fedpseudo}.
Since this field is in its early development, reproducible and standardized simulation environments are paramount for the comparability of results.
Simulation environments mimic one or many aspects of real-world federations, such as client availability, communication constraints, computation constraints, and data heterogeneity.
Some existing works provide simulation environments for standard federated learning applications~\cite{tff,beutel2020flower}.
However, these environments have no direct support for survival analysis problems.
Other works implement algorithms for non-federated survival models~\cite{sksurv,kvamme2019time,pysurvival_cite,davidson2019lifelines} based on centralized survival datasets~\cite{drysdale2022survset}.
Recently, a benchmarking suite for federated healthcare problems has been developed, including a single federated survival dataset with a predefined data split among 6 clients~\cite{terrail2022flamby}.

The goal of this work is to extend the benchmarking ground for federated survival models.
To this end, we present a novel technique for constructing realistic federated datasets from existing non-federated survival datasets in a flexible and reproducible way.
Realistic federated datasets mimic real-world heterogeneity by exhibiting non-identically distributed (non-IID) data among clients. 
More specifically, we provide two algorithms for assigning each data sample from a centralized survival dataset to a carefully chosen client.
The proposed algorithms are based on the Dirichlet distribution~\cite{hsu2019measuring,li2022federated}, as it can induce distribution skewness by tuning a single parameter.
The first algorithm focuses on building federated datasets with a non-uniform number of samples.
We call this algorithm quantity-skewed splitting.
The second one, instead, builds client datasets with different label distributions.
We call this algorithm label-skewed splitting.
The heterogeneity level introduced by each algorithm in the resulting data assignments can be tuned with a parameter $\alpha > 0$, such that for $\alpha \rightarrow 0$ data are more skewed, while for $\alpha \rightarrow \infty$ data are more uniform.
The ability to tune the heterogeneity level allows for federated simulations with different environmental conditions.
This aspect is essential to test the resilience of federated survival models to non-IID realistic data distributions.

The presented techniques have been tested on a collection of datasets for survival analysis, providing visual insights about the level of heterogeneity induced in each setting.
Also, the level of heterogeneity is numerically investigated with log-rank tests~\cite{bland2004logrank} within client distributions.
The experimental evaluation demonstrates that the proposed techniques are able to build heterogeneous federated datasets starting from non-federated survival data.
Moreover, the numerical analysis shows how the $\alpha$ parameter can effectively control the heterogeneity level induced by each split.

The implementation of quantity-skewed and label-skewed splitting is publicly available~\cite{archetti2023heterogeneous} in favor of reproducibility and to encourage the usage of common practices in the simulation of federated survival environments.

%% file: src/2_related.tex
\section{Background and Related Works}

This section summarizes the main aspects of survival analysis and federated learning and reviews the state-of-the-art on federated survival analysis.

\subsection{Survival Analysis}

Survival analysis, also known as time-to-event analysis, is a statistical machine learning field that models the occurrence time of an event of interest for a population~\cite{wang2019machine}.
The distinctive feature of survival models is the handling of censored data.
With censored data, we refer to samples for which the event occurrence was not observed during the study.
A survival dataset $D$ is a set of $N$ triplets 
\begin{equation*}
    (\mathbf{x}_i, \delta_i, t_i)\text{, } i = 1, \dots, N \text{ s.t.}
\end{equation*}
\begin{itemize}
    \item $\mathbf{x}_i \in \mathbb{R}^d$ is a $d$-dimensional feature vector, also called covariate vector, that retains all the input information for a sample;
    \item $\delta_i$ is the event occurrence indicator. If $\delta_i = 1$, then the $i$-th sample experienced the event, otherwise the $i$-th sample is censored and $\delta_i = 0$;
    \item $t_i=\min \left\{t_i^e, t_i^c\right\}$ is the minimum between the actual event time $t_i^e$ and the censoring time $t_i^c$.
\end{itemize}
This setting refers to right-censoring~\cite{lee2003statistical}, where the censoring time is less than or equal to the actual event time.
This is the case, for instance, of disease recurrence under a certain treatment~\cite{schumacher1994randomized} or patient death~\cite{dispenzieri2012use}.
Indeed, right-censoring is the most common scenario in real-world survival applications~\cite{wang2019machine}.
Therefore, we limit the discussion to the right censoring setting for the rest of the paper.

The goal of survival analysis is to estimate the event occurrence probability with respect to time.
In particular, the output of a survival model is the survival function
\begin{equation*}
    S(t | \mathbf{x}) = P(T > t | \mathbf{x}).
\end{equation*}

Survival models are classified into three types: non-parametric, semi-parametric, and parametric~\cite{wang2019machine}.
In this work, we include non-parametric models in the analysis of the proposed data splitting algorithms, as these are the only models that make no assumption about the underlying event distribution over time.
Moreover, non-parametric models are well-suited for survival data visualization.
Indeed, non-parametric models encode the overall survival behavior of a population by predicting a survival function $\hat{S}(t)$ which is not conditioned on $\mathbf{x}$.

Non-parametric models are Kaplan-Meier (KM)~\cite{kaplan1958nonparametric}, Nelson-Aalen~\cite{nelson1972theory,aalen1978nonparametric}, and Life-Table~\cite{cutler1958maximum}.
Among those, the KM estimator is the most widely spread in survival applications due to its intuitive interpretation.
The KM estimator starts from the set of unique event occurrence times $T_D = \{t_j: (\mathbf{x}_i,\delta_i,t_j) \in D\}$. 
Then, for each $t_j \in T_D$ it computes the number of observed events $d_j \geq 1$ at time $t_j$ and the number of samples $r_j$ that did not yet experience an event.
The KM estimator is computed as 
\begin{equation*}
    \hat{S}(t) = \prod\limits_{j: t_j < t} \left(1 - \frac{d_j}{r_j} \right).
\end{equation*}

\subsection{Federated Learning}

Federated Learning (FL)~\cite{li2020federated,kairouz2021advances} is a machine learning setting in which a set of agents jointly train a model without sharing the data they store locally.
FL algorithms rely on a central server for message exchange and agent coordination.
A federation is composed of $K$ clients, each holding a private dataset $D_k$, $k = 1, \dots, K$.
The goal of a FL algorithm is to find the best parameters $w$ that optimize a global loss function $\mathcal{L}$:
\begin{equation*}
    \min_w \mathcal{L}(w) = \min_w\sum\limits_{k=1}^K \lambda_k \mathcal{L}_k(w).
    \label{eq:fl}
\end{equation*}
$\mathcal{L}_k$ is the local loss function computed by client $k$. 
$\lambda_k$ is a set of parameters weighting the contribution of each client to the global loss.
Usually, $\lambda_k$ is proportional to the number of samples on which each client $k$ evaluated $\mathcal{L}_k(w)$ locally.
This weighting strategy favors contributions from clients holding more private data, which are more likely to be representative of the entire data distribution.

Federated Averaging (FedAvg)~\cite{mcmahan2017communication} is the first algorithm developed to minimize $\mathcal{L}$.
It relies on iterative averaging of model parameters trained locally on random subsets of clients.
However, FedAvg is not always suited to face system security and confidentiality preservation challenges in real-world applications~\cite{mothukuri2021survey,rahimian2022practical}.
Moreover, real-world applications present multiple levels of heterogeneity.
First, system heterogeneity constraints FL algorithms to comply with the hardware limitations of the network channel and the clients'~devices.
Second, datasets are not guaranteed to contain identically distributed data.
In fact, in most real-world scenarios data are likely to be non-IID.
In order to handle data heterogeneity in federated environments, several non-survival federated algorithms have been proposed~\cite{li2020fedprox,karimireddy2020scaffold,acar2021federated}.

\subsection{Federated Survival Analysis}

Federated learning provides key advantages for the future of healthcare applications~\cite{rieke2020future}.
In particular, federated survival analysis investigates the opportunities and challenges related to the integration of federated learning into survival analysis tasks.
However, few works specifically tackle federated survival analysis applications.
Some works~\cite{andreux2020federated,wang2022survmaximin} provide solutions for the non-separability of the partial log-likelihood loss, used to train Cox survival models~\cite{cox1972regression}.
Indeed, non-separable loss functions are not suited for federated learning algorithms, as their evaluation requires access to all the available data in the federation.
Other works~\cite{lu2015webdisco,rahman2022fedpseudo} provide federated versions of classical survival algorithms asymptotically equivalent to their centralized counterparts.
Within these works, data federations are built with uniform data splits or with entirely simulated datasets.

\subsection{Federated Datasets}

Concerning the available datasets for federated simulation, LEAF~\cite{caldas2018leaf} is the most widely spread dataset collection for standard federated learning applications.
It provides several real-world datasets covering classification, sentiment analysis, next-character, and next-word prediction.
Secure Generative Data Exchange (SGDE)~\cite{lomurno2022sgde} is a recent framework to build synthetic datasets in a privacy-preserving way.
SGDE provides inherently heterogeneous datasets composed of synthetic samples provided by client-side data generators.
Currently, SGDE has been applied to classification and regression problems only.
Other studies~\cite{hsu2019measuring,li2022federated} investigate the taxonomy of data heterogeneity and provide techniques to emulate non-IID data splits starting from centralized classification datasets.
Recently, FLamby~\cite{terrail2022flamby} provided a set of benchmarking tools for cross-silo federated applications concerning healthcare. 
Among those, Fed-TCGA-BRCA is a federated survival dataset collecting the data of 1066 patients geographically grouped into 6 clients.

To the best of our knowledge, Fed-TCGA-BRCA is the only federated survival dataset proposed to date.
Moreover, existing data-splitting techniques are tailored for non-survival problems only.
This is the first study extending data-splitting techniques to survival analysis, providing greater flexibility for modeling simulated survival environments.

%% file: src/3_method.tex
\section{Method}

This paper presents a set of techniques to split survival datasets into heterogeneous federations.
We start from a survival dataset $D$ and a number of clients $K$. 
The goal is to assign to each sample in $D$ a client $k \in \{1, \dots, K\}$, such that federated survival algorithms can leverage the set of $D_k$s to simulate heterogeneous learning scenarios.
The work proposes two splitting techniques: quantity-skewed and label-skewed splitting.

\subsection{Quantity-Skewed Splitting}
\label{sec:qss}

Quantity-skewed splitting pertains to a scenario where the number of samples for each client $k$, represented as $|D_k|$, varies among clients. 
In such a scenario, clients with a limited number of samples may generate gradients that are inherently noisy, which can impede the convergence of federated learning algorithms. 
This is due to the fact that clients with a smaller number of samples tend to exhibit higher variance in their gradients, leading to instability in the federated learning process and hampering convergence rate.

Simulation of quantity-skewed scenarios is essential in assessing the robustness of federated survival algorithms. 
It enables researchers to evaluate the algorithm's ability to handle the imbalance in sample distribution across clients and its impact on algorithm performance.

Similarly to~\cite{hsu2019measuring,li2022federated}, the proportion of samples $\mathbf{p}$ to assign to each client follows a Dirichlet distribution
\begin{equation*}
    \mathbf{p} \sim \mathcal{D}(\alpha \cdot \mathbf{1}_K).
\end{equation*}
Here, $\mathbf{1}_K$ is a vector of $1$s of length $K$.
$\mathbf{p} \in [0, 1]^K$ such that $\langle \mathbf{1}_K, \mathbf{p}\rangle = 1$.
$\alpha > 0$ is a similarity parameter controlling the similarity between client dataset cardinalities $|D_k|$.
For $\alpha \rightarrow 0$, the number of samples for each $D_k$ are heterogeneous.
Conversely, for $\alpha \rightarrow \infty$, the number of samples for each $D_k$ tends to be similar.
With quantity-skewed splitting, each sample $(\mathbf{x}_i,\delta_i,t_i)$ is assigned to a client dataset $D_k$ with probability 
\begin{equation*}
    P\left((\mathbf{x}_i,\delta_i,t_i\right) \in D_k) = \mathbf{p}[k].
\end{equation*}

\subsection{Label-Skewed Splitting}
\label{sec:lss}

Label-skewed splitting pertains to scenarios in which the distribution of labels differs among client datasets. 
This type of distribution heterogeneity is commonly encountered in real-world federated learning scenarios. 
The non-IID distribution can be attributed to various factors, including variations in data collection and storage processes, the use of different acquisition devices, and variations in preprocessing or labeling techniques. 
Additionally, clients may have different label quantities due to domain-specific factors. 
For instance, in a federated healthcare scenario for treatment risk assessment, one client may have a dataset of records from a rural hospital, while another client may have data from an urban hospital. 
These datasets from different locations may exhibit heterogeneous label distributions due to disparities in patient demographics and healthcare access.

To produce a label-skewed data split, first, the timeline of the original survival dataset is divided into $B$ bins, obtaining a set of time instants $\{\tau_0, \dots, \tau_{B}\}$.
The bin identification can be uniform or quantile-based, as in~\cite{kvamme2021continuous}.
Then, each sample $(\mathbf{x}_i,\delta_i,t_i)$ is assigned a class that corresponds to the $b$-th bin, such that $t_i \in (\tau_{b-1},\tau_{b}]$.
Following~\cite{hsu2019measuring,li2022federated}, the Dirichlet distribution is used to identify heterogeneous splitting proportions according to the sample class as
\begin{equation*}
    \arraycolsep=1.4pt
    \begin{array}{@{}cll@{}}
        \arraycolsep=1.4pt\def\arraystretch{2.2}
        \mathbf{p}_1 & \sim & \mathcal{D}(\alpha \cdot \mathbf{1}_K) \\
        \vdots && \\
        \mathbf{p}_B & \sim & \mathcal{D}(\alpha \cdot \mathbf{1}_K) \\
    \end{array}
\end{equation*}
Finally, each sample $(\mathbf{x}_i,\delta_i,t_i)$ assigned to label $b$ is added to $D_k$ with probability 
\begin{equation*}
    P\left((\mathbf{x}_i,\delta_i,t_i\right) \in D_k) = \mathbf{p}_b[k].
\end{equation*}
The $\alpha$ parameter controls the level of similarity between label distributions.
For $\alpha \rightarrow \infty$, client label distributions are similar, while for $\alpha \rightarrow 0$ label distributions differ.
The numerical dependency between $\alpha$ and the data heterogeneity level is discussed in detail using log-rank tests~\cite{bland2004logrank} in Section~\ref{sec:experiments}.

%% file: src/4_experiments.tex
\section{Experiments}
\label{sec:experiments}

This section presents the experiments carried out to evaluate the proposed methods for building heterogeneous datasets for federated survival analysis.

\subsection{Datasets}
\label{sec:datasets}

\input{tab/datasets.tex}

Each of the experiments involves the following survival datasets:
the German Breast Cancer Study Group 2 (GBSG2)~\cite{schumacher1994randomized}, 
the Molecular Taxonomy of Breast Cancer International Consortium (METABRIC)~\cite{katzman2018deepsurv}, 
the Australian AIDS survival dataset (AIDS)~\cite{ripley2023mass}, 
the assay of serum-free light chain dataset (FLCHAIN)~\cite{therneau2023survival}, and 
the Study to Understand Prognoses Preferences Outcomes and Risks of Treatment (SUPPORT)~\cite{vanderbilt2022vanderbilt}.
The dataset summary statistics are collected in Table~\ref{tab:survival-datasets}.

\subsection{Visualizing Splitting Methods}

\begin{figure*}
    \centering
    \includegraphics[width=.96\textwidth]{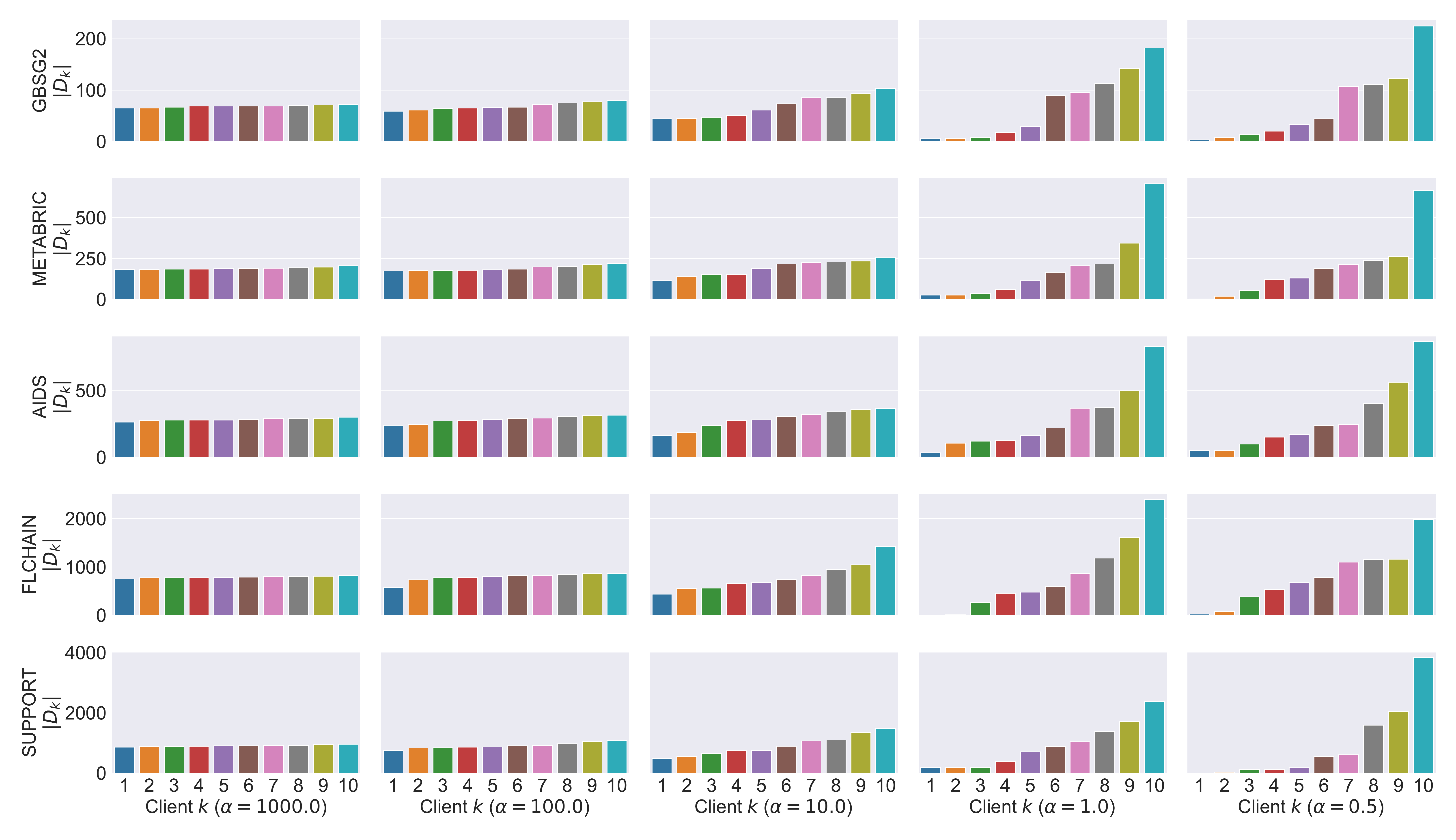}
    \caption{Number of samples $|D_k|$ for each client $k=1, \dots, 10$. Each row refers to one of the datasets described in Section~\ref{sec:datasets}. Each column corresponds to a quantity-skewed split (Section~\ref{sec:qss}) with a fixed similarity parameter $\alpha$.}
    \label{fig:us}
\end{figure*}

\begin{figure*}
    \centering
    \includegraphics[width=.96\textwidth]{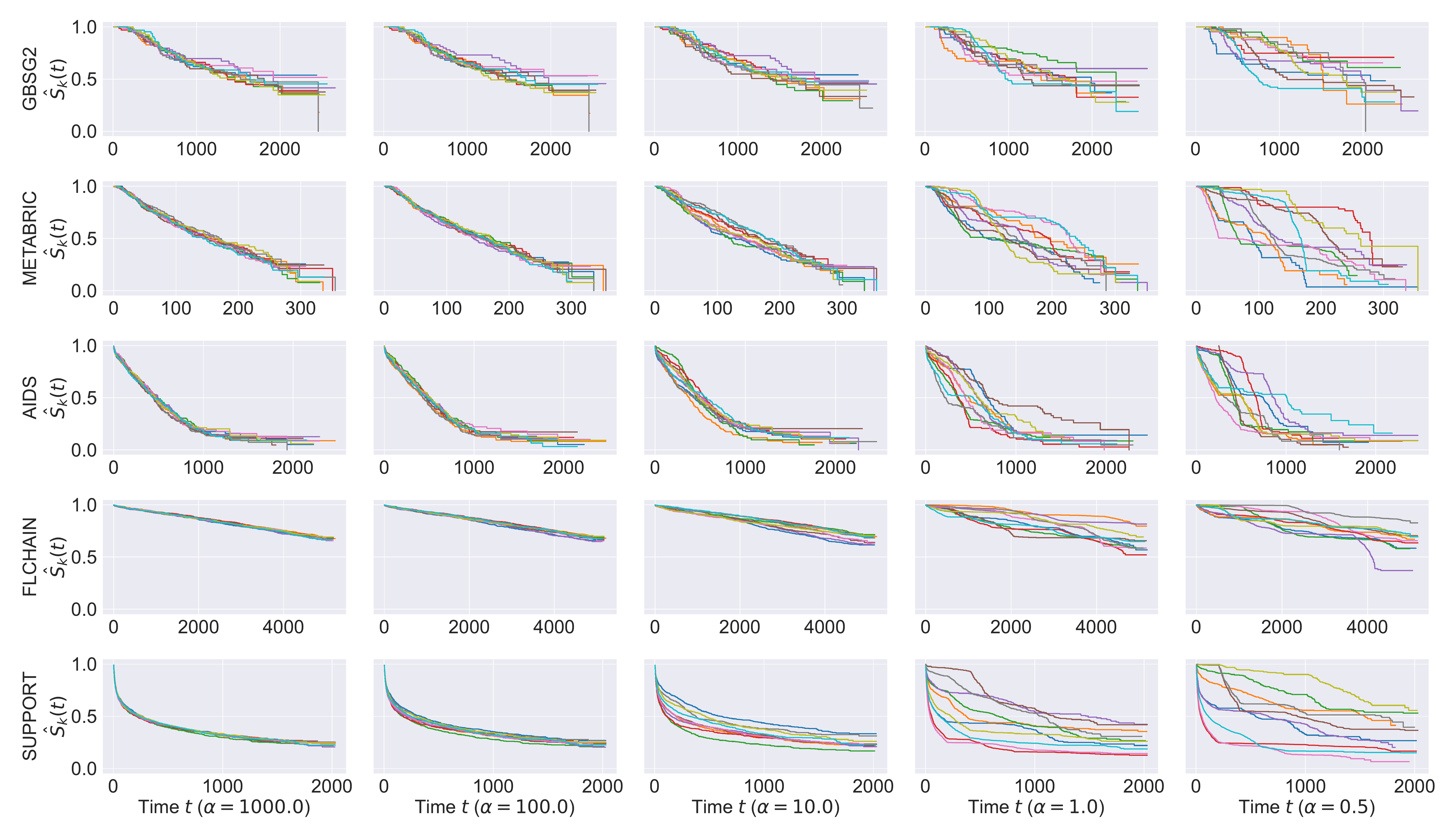}
    \caption{Kaplan-Meier estimators $\hat{S}_k(t)$ for each client $k=1, \dots, 10$. Each row refers to one of the datasets described in Section~\ref{sec:datasets}. Each column corresponds to a label-skewed split (Section~\ref{sec:lss}) with a fixed similarity parameter $\alpha$.}
    \label{fig:ls}
\end{figure*}

This section describes the visual results obtained from the splitting methods under different $\alpha$ parameters.
In particular, Figure~\ref{fig:us} shows the results of the quantity-skewed splitting algorithm described in Section~\ref{sec:qss}.
Splits are generated for a federation of 10 clients ($K=10$), each identified by a different color in the plots.
In Figure~\ref{fig:us}, each row corresponds to one of the example datasets described in Section~\ref{sec:datasets}.
Columns, instead, refer to different values of the similarity parameter $\alpha$, ranging from $\alpha = 1000$ (low heterogeneity) to $\alpha = 0.5$ (high heterogeneity).
Each plot shows the client dataset cardinalities $|D_k|$ with respect to clients $k=1, \dots, 10$.
By looking at the plots on the left of Figure~\ref{fig:us}, higher values of $\alpha$ tend to produce similar dataset cardinalities $|D_k|$.
Conversely, for lower $\alpha$ values, $|D_k|$s considerably differ between clients.
This trend is confirmed for all the datasets involved in the experiments.

Figure~\ref{fig:ls} shows the results of the label-skewed splitting algorithm described in Section~\ref{sec:lss}.
Each plot shows the Kaplan-Meier estimators $\hat{S}_k(t)$ of each client dataset $D_k$, $k=1, \dots, 10$.
The KM estimator an excellent tool for survival function visualization, as it encodes the summary information concerning the survival labels in the dataset.
Following the structure of Figure~\ref{fig:us}, in Figure~\ref{fig:ls} each row corresponds to a specific dataset from Section~\ref{sec:datasets} and each column corresponds to an $\alpha$ value, decreasing from $1000$ to $0.5$. 
From the left column to the right column, the KM estimators of each client tend to separate, as $\alpha$ decreases.
This is expected, as for lower $\alpha$ values, the Dirichlet distribution assigns non-uniform proportions of samples from each time bin to the clients.
In this way, decreasing the $\alpha$ parameter results in heterogeneous federated distributions.

\subsection{Numerical Analysis of Heterogeneity}

\input{tab/het_level.tex}

This section provides the quantitative analysis carried out to evaluate the level of heterogeneity induced by each splitting method.
A high level of data heterogeneity entails different client data distributions, which leads to more realistic federations.
We use the log-rank test~\cite{bland2004logrank} to determine whether the event occurrence distribution is the same for two clients.
This test verifies the null hypothesis that there is no statistically significant difference between the survival distributions of two given populations.
We consider the distribution difference between two clients $k_1, k_2$ statistically significant if the p-value $p_{k_1,k_2}$ resulting from the test is $\leq 0.05$.

In order to summarize the results for a federation, we define the heterogeneity score $h$ of a federation as the fraction of client pairs $\mathcal{P} = \{(k_1,k_2: k_1 < k_2 \wedge k_1,k_2=1, \dots, K)\}$ for which the distribution difference is statistically significant, i.e.,
\begin{equation*}
    h = \frac{1}{|\mathcal{P}|}\sum\limits_{(k_1,k_2) \in \mathcal{P}} \mathbb{1}(p_{k_1,k_2} \leq 0.05).
\end{equation*}
Table~\ref{tab:logrank} collects the $h$ values for quantity-skewed and label-skewed splits under several $K$ and $\alpha$ values.
Each result is averaged over 100 runs.

Concerning quantity-skewed splitting, each setting presents an average heterogeneity score smaller than 5\%.
In other words, quantity-skewed survival data does not present statistically significant label distribution differences when comparing pairs of client datasets.
This implies that quantity-skewed splitting is well suited to test the resilience of a federated algorithm to data imbalance, but not to heterogeneous data distributions.

Conversely, label-skewed splitting exhibits noticeable differences in $h$ scores depending on the value of $\alpha$. 
In fact, for all the tested datasets, the $h$ score with $\alpha = 1000$ is almost zero, and decreasing $\alpha$ affects the number of different label distributions among clients.
For datasets with smaller total cardinalities (GBSG2, METABRIC, and AIDS) $\alpha$ must be smaller than $10$ in order to detect noticeable differences between client distributions.
Instead, datasets with more total samples (FLCHAIN and SUPPORT) present high heterogeneity even for $\alpha = 100$.
For all the dataset splits in small federations ($K = 5$ and $K = 10$), $\alpha$ values smaller than $1$ result in $h > 50$\%.
The trend does not apply to federations with more clients ($K=50$), where even $\alpha = 0.1$ is not enough to obtain $h > 50$\%.

%% file: tab/datasets.tex
\begin{table}[t]
    \centering
    \caption{Survival datasets involved in the experiments.}
    \label{tab:survival-datasets}
    
    \begin{tabularx}{\linewidth}{@{} X Y Y Y @{}}
        \toprule
        \textbf{Dataset} & \textbf{Samples} & \textbf{Censored} & \textbf{Features}\\
        \midrule
        GBSG~\cite{schumacher1994randomized} & 686 & 44\% & 8\\
        METABRIC~\cite{katzman2018deepsurv} & 1904 & 58\% & 8\\
        AIDS~\cite{ripley2023mass} & 2839 & 62\% & 4\\
        FLCHAIN~\cite{therneau2023survival} & 7874 & 28\% & 10\\
        SUPPORT~\cite{vanderbilt2022vanderbilt} & 9105 & 68\% & 35\\
        \bottomrule
    \end{tabularx}
\end{table}

%% file: tab/het_level.tex
\begin{table*}[t]
\centering
\caption{Heterogeneity score $h$ for several $K$ and $\alpha$. $h$ values are averaged over 100 runs and scaled by 100 for better readability.}
\label{tab:logrank}
\begin{tabularx}{\linewidth}{@{} X Y Y Y Y Y Y @{}}
\toprule

\multicolumn{7}{c}{Quantity-Skewed Split, $K = 5$} \\
\midrule
Dataset & $\alpha = 1000.0$ & $\alpha = 100.0$ & $\alpha = 10.0$ & $\alpha = 1.0$ & $\alpha = 0.5$ & $\alpha = 0.1$ \\
\midrule
GBSG2 & 2.6$\pm$6.0 & 3.1$\pm$7.1 & 3.4$\pm$8.2 & 2.1$\pm$5.9 & 4.3$\pm$9.3 & 2.2$\pm$6.6\\
METABRIC & 2.8$\pm$7.3 & 3.3$\pm$7.9 & 3.1$\pm$7.6 & 2.9$\pm$8.3 & 1.5$\pm$5.4 & 2.2$\pm$7.5\\
AIDS & 1.4$\pm$5.3 & 2.8$\pm$6.5 & 2.1$\pm$5.0 & 4.6$\pm$10.5 & 4.6$\pm$9.8 & 2.3$\pm$5.8\\
FLCHAIN & 1.9$\pm$4.6 & 3.2$\pm$6.9 & 2.3$\pm$6.0 & 3.8$\pm$8.4 & 2.9$\pm$9.8 & 2.6$\pm$6.8\\
SUPPORT & 3.0$\pm$6.9 & 2.0$\pm$4.7 & 2.5$\pm$6.7 & 3.3$\pm$7.4 & 3.7$\pm$9.4 & 0.3$\pm$2.2\\
\midrule
\multicolumn{7}{c}{Quantity-Skewed Split, $K = 10$} \\
\midrule
Dataset & $\alpha = 1000.0$ & $\alpha = 100.0$ & $\alpha = 10.0$ & $\alpha = 1.0$ & $\alpha = 0.5$ & $\alpha = 0.1$ \\
\midrule
GBSG2 & 4.1$\pm$4.8 & 3.9$\pm$5.0 & 3.0$\pm$4.9 & 3.0$\pm$4.3 & 3.0$\pm$4.5 & 1.9$\pm$3.3\\
METABRIC & 3.6$\pm$5.3 & 4.7$\pm$6.0 & 4.4$\pm$5.9 & 3.5$\pm$5.6 & 3.6$\pm$4.7 & 1.7$\pm$4.0\\
AIDS & 4.1$\pm$5.6 & 4.5$\pm$5.5 & 3.8$\pm$5.4 & 4.5$\pm$6.4 & 4.5$\pm$6.3 & 2.3$\pm$3.6\\
FLCHAIN & 3.7$\pm$4.8 & 3.4$\pm$5.0 & 3.6$\pm$4.5 & 5.5$\pm$6.7 & 4.2$\pm$6.2 & 2.3$\pm$4.0\\
SUPPORT & 4.1$\pm$5.8 & 3.4$\pm$4.6 & 4.0$\pm$4.8 & 3.9$\pm$5.7 & 4.2$\pm$6.5 & 1.0$\pm$2.3\\
\midrule
\multicolumn{7}{c}{Quantity-Skewed Split, $K = 50$} \\
\midrule
Dataset & $\alpha = 1000.0$ & $\alpha = 100.0$ & $\alpha = 10.0$ & $\alpha = 1.0$ & $\alpha = 0.5$ & $\alpha = 0.1$ \\
\midrule
GBSG2 & 3.9$\pm$2.0 & 3.4$\pm$1.8 & 3.5$\pm$2.0 & 3.0$\pm$1.8 & 2.6$\pm$1.7 & 1.6$\pm$1.0\\
METABRIC & 4.6$\pm$2.3 & 4.7$\pm$2.6 & 4.4$\pm$2.0 & 3.9$\pm$2.1 & 3.2$\pm$1.8 & 1.5$\pm$1.1\\
AIDS & 4.5$\pm$2.2 & 4.9$\pm$2.5 & 4.4$\pm$2.1 & 4.6$\pm$2.4 & 4.2$\pm$2.4 & 2.0$\pm$1.1\\
FLCHAIN & 4.8$\pm$2.4 & 5.0$\pm$2.4 & 4.6$\pm$2.2 & 4.7$\pm$2.6 & 4.8$\pm$2.7 & 2.1$\pm$1.5\\
SUPPORT & 4.5$\pm$2.2 & 4.5$\pm$2.3 & 4.8$\pm$2.3 & 3.9$\pm$2.1 & 3.4$\pm$2.1 & 0.8$\pm$0.8\\
\midrule
\midrule
\multicolumn{7}{c}{Label-Skewed Split, $K = 5$} \\
\midrule
Dataset & $\alpha = 1000.0$ & $\alpha = 100.0$ & $\alpha = 10.0$ & $\alpha = 1.0$ & $\alpha = 0.5$ & $\alpha = 0.1$ \\
\midrule
GBSG2 & 0.2$\pm$2.0 & 0.1$\pm$1.0 & 5.8$\pm$9.4 & 46.7$\pm$20.9 & 58.2$\pm$17.0 & 73.8$\pm$18.2\\
METABRIC & 0.0$\pm$0.0 & 0.5$\pm$2.2 & 20.9$\pm$17.2 & 66.1$\pm$19.0 & 76.7$\pm$14.5 & 82.3$\pm$13.3\\
AIDS & 0.3$\pm$1.7 & 3.1$\pm$7.2 & 37.5$\pm$21.9 & 75.1$\pm$16.2 & 81.5$\pm$14.4 & 86.6$\pm$11.3\\
FLCHAIN & 0.3$\pm$1.7 & 12.6$\pm$14.9 & 58.8$\pm$17.6 & 83.9$\pm$12.4 & 88.0$\pm$11.4 & 94.1$\pm$7.0\\
SUPPORT & 0.5$\pm$2.2 & 29.6$\pm$20.8 & 74.3$\pm$15.7 & 91.3$\pm$9.7 & 92.5$\pm$7.4 & 94.0$\pm$6.4\\
\midrule
\multicolumn{7}{c}{Label-Skewed Split, $K = 10$} \\
\midrule
Dataset & $\alpha = 1000.0$ & $\alpha = 100.0$ & $\alpha = 10.0$ & $\alpha = 1.0$ & $\alpha = 0.5$ & $\alpha = 0.1$ \\
\midrule
GBSG2 & 0.4$\pm$1.5 & 0.6$\pm$1.5 & 2.8$\pm$4.3 & 32.2$\pm$11.7 & 43.7$\pm$11.5 & 63.2$\pm$12.9\\
METABRIC & 0.1$\pm$0.4 & 0.2$\pm$1.0 & 10.6$\pm$8.4 & 54.6$\pm$13.6 & 66.5$\pm$10.1 & 76.7$\pm$8.7\\
AIDS & 0.3$\pm$1.0 & 1.4$\pm$2.7 & 24.7$\pm$12.6 & 68.1$\pm$9.0 & 74.0$\pm$9.1 & 77.7$\pm$8.4\\
FLCHAIN & 0.4$\pm$1.2 & 4.2$\pm$5.5 & 42.8$\pm$13.0 & 78.2$\pm$8.8 & 84.9$\pm$5.6 & 89.3$\pm$5.8\\
SUPPORT & 0.1$\pm$0.4 & 14.7$\pm$9.7 & 63.2$\pm$10.5 & 87.0$\pm$4.9 & 88.5$\pm$4.7 & 89.7$\pm$6.1\\
\midrule
\multicolumn{7}{c}{Label-Skewed Split, $K = 50$} \\
\midrule
Dataset & $\alpha = 1000.0$ & $\alpha = 100.0$ & $\alpha = 10.0$ & $\alpha = 1.0$ & $\alpha = 0.5$ & $\alpha = 0.1$ \\
\midrule
GBSG2 & 0.5$\pm$0.6 & 0.6$\pm$0.6 & 0.5$\pm$0.6 & 5.7$\pm$2.2 & 10.8$\pm$3.3 & 23.8$\pm$4.2\\
METABRIC & 0.2$\pm$0.3 & 0.3$\pm$0.5 & 1.3$\pm$1.2 & 21.7$\pm$4.5 & 33.1$\pm$5.2 & 48.8$\pm$5.5\\
AIDS & 0.6$\pm$0.5 & 0.8$\pm$0.8 & 4.5$\pm$2.3 & 34.6$\pm$5.2 & 45.3$\pm$4.4 & 49.8$\pm$4.9\\
FLCHAIN & 0.2$\pm$0.3 & 0.6$\pm$0.6 & 10.6$\pm$3.7 & 55.4$\pm$4.1 & 64.8$\pm$2.6 & 72.4$\pm$3.6\\
SUPPORT & 0.0$\pm$0.0 & 0.5$\pm$0.6 & 29.0$\pm$5.5 & 69.9$\pm$2.7 & 75.5$\pm$2.3 & 73.4$\pm$4.2\\
\bottomrule

\end{tabularx}
\end{table*}

%% file: src/5_conclusion.tex
\section{Conclusion}

This paper proposes two algorithms to simulate data heterogeneity in survival datasets for federated learning.
Federated simulation is an important step in survival analysis toward the implementation and production of more accurate, fair, and privacy-preserving survival models.
The presented splitting techniques are based on the Dirichlet distribution.
Quantity-skewed splitting produces datasets with variable cardinalities, while label-skewed splitting relies on time binning to split samples according to different label distributions.
Visual insights are provided to show the behavior of the proposed methods under hyperparameter change.
Moreover, log-rank tests are reported to provide a quantitative evaluation of the degree of heterogeneity induced by each data split.
To encourage the adoption of common benchmarking practices for future experiments on federated survival analysis, we make the source code of the proposed algorithms publicly available.

%% file: main.bbl
\begin{thebibliography}{10}
\providecommand{\url}[1]{#1}
\csname url@samestyle\endcsname
\providecommand{\newblock}{\relax}
\providecommand{\bibinfo}[2]{#2}
\providecommand{\BIBentrySTDinterwordspacing}{\spaceskip=0pt\relax}
\providecommand{\BIBentryALTinterwordstretchfactor}{4}
\providecommand{\BIBentryALTinterwordspacing}{\spaceskip=\fontdimen2\font plus
\BIBentryALTinterwordstretchfactor\fontdimen3\font minus
  \fontdimen4\font\relax}
\providecommand{\BIBforeignlanguage}[2]{{%
\expandafter\ifx\csname l@#1\endcsname\relax
\typeout{** WARNING: IEEEtran.bst: No hyphenation pattern has been}%
\typeout{** loaded for the language `#1'. Using the pattern for}%
\typeout{** the default language instead.}%
\else
\language=\csname l@#1\endcsname
\fi
#2}}
\providecommand{\BIBdecl}{\relax}
\BIBdecl

\bibitem{klein2003survival}
J.~P. Klein and M.~L. Moeschberger, \emph{Survival analysis: techniques for
  censored and truncated data}.\hskip 1em plus 0.5em minus 0.4em\relax
  Springer, 2003, vol. 1230.

\bibitem{wang2019machine}
P.~Wang, Y.~Li, and C.~K. Reddy, ``Machine learning for survival analysis: A
  survey,'' \emph{ACM Computing Surveys (CSUR)}, vol.~51, no.~6, pp. 1--36,
  2019.

\bibitem{emmert2019introduction}
F.~Emmert-Streib and M.~Dehmer, ``Introduction to survival analysis in
  practice,'' \emph{Machine Learning and Knowledge Extraction}, vol.~1, no.~3,
  pp. 1013--1038, 2019.

\bibitem{andreux2020federated}
M.~Andreux, A.~Manoel, R.~Menuet, C.~Saillard, and C.~Simpson, ``Federated
  survival analysis with discrete-time cox models,'' \emph{arXiv preprint
  arXiv:2006.08997}, 2020.

\bibitem{rieke2020future}
N.~Rieke, J.~Hancox, W.~Li, F.~Milletari, H.~R. Roth, S.~Albarqouni, S.~Bakas,
  M.~N. Galtier, B.~A. Landman, K.~Maier-Hein \emph{et~al.}, ``The future of
  digital health with federated learning,'' \emph{NPJ digital medicine},
  vol.~3, no.~1, pp. 1--7, 2020.

\bibitem{li2020federated}
T.~Li, A.~K. Sahu, A.~Talwalkar, and V.~Smith, ``Federated learning:
  Challenges, methods, and future directions,'' \emph{IEEE Signal Processing
  Magazine}, vol.~37, no.~3, pp. 50--60, 2020.

\bibitem{kairouz2021advances}
P.~Kairouz, H.~B. McMahan, B.~Avent, A.~Bellet, M.~Bennis, A.~N. Bhagoji,
  K.~Bonawitz, Z.~Charles, G.~Cormode, R.~Cummings \emph{et~al.}, ``Advances
  and open problems in federated learning,'' \emph{Foundations and Trends in
  Machine Learning}, vol.~14, no. 1--2, pp. 1--210, 2021.

\bibitem{wang2022survmaximin}
X.~Wang, H.~G. Zhang, X.~Xiong, C.~Hong, G.~M. Weber, G.~A. Brat, C.-L. Bonzel,
  Y.~Luo, R.~Duan, N.~P. Palmer \emph{et~al.}, ``Survmaximin: robust federated
  approach to transporting survival risk prediction models,'' \emph{Journal of
  biomedical informatics}, vol. 134, p. 104176, 2022.

\bibitem{lu2015webdisco}
C.-L. Lu, S.~Wang, Z.~Ji, Y.~Wu, L.~Xiong, X.~Jiang, and L.~Ohno-Machado,
  ``Webdisco: a web service for distributed cox model learning without
  patient-level data sharing,'' \emph{Journal of the American Medical
  Informatics Association}, vol.~22, no.~6, pp. 1212--1219, 2015.

\bibitem{rahman2022fedpseudo}
M.~M. Rahman and S.~Purushotham, ``Fedpseudo: Pseudo value-based deep learning
  models for federated survival analysis,'' \emph{arXiv preprint
  arXiv:2207.05247}, 2022.

\bibitem{tff}
\BIBentryALTinterwordspacing
T.~T.~F. Authors, ``{TensorFlow Federated},'' 12 2018. [Online]. Available:
  \url{https://github.com/tensorflow/federated}
\BIBentrySTDinterwordspacing

\bibitem{beutel2020flower}
D.~J. Beutel, T.~Topal, A.~Mathur, X.~Qiu, T.~Parcollet, P.~P. de~Gusm{\~a}o,
  and N.~D. Lane, ``Flower: A friendly federated learning research framework,''
  \emph{arXiv preprint arXiv:2007.14390}, 2020.

\bibitem{sksurv}
\BIBentryALTinterwordspacing
S.~P{\"o}lsterl, ``scikit-survival: A library for time-to-event analysis built
  on top of scikit-learn,'' \emph{Journal of Machine Learning Research},
  vol.~21, no. 212, pp. 1--6, 2020. [Online]. Available:
  \url{http://jmlr.org/papers/v21/20-729.html}
\BIBentrySTDinterwordspacing

\bibitem{kvamme2019time}
H.~Kvamme, {\O}.~Borgan, and I.~Scheel, ``Time-to-event prediction with neural
  networks and cox regression,'' \emph{arXiv preprint arXiv:1907.00825}, 2019.

\bibitem{pysurvival_cite}
\BIBentryALTinterwordspacing
S.~Fotso \emph{et~al.}, ``{PySurvival}: Open source package for survival
  analysis modeling,'' 2019--. [Online]. Available:
  \url{https://www.pysurvival.io/}
\BIBentrySTDinterwordspacing

\bibitem{davidson2019lifelines}
C.~Davidson-Pilon, ``lifelines: survival analysis in python,'' \emph{Journal of
  Open Source Software}, vol.~4, no.~40, p. 1317, 2019.

\bibitem{drysdale2022survset}
E.~Drysdale, ``Survset: An open-source time-to-event dataset repository,''
  \emph{arXiv preprint arXiv:2203.03094}, 2022.

\bibitem{terrail2022flamby}
J.~O.~d. Terrail, S.-S. Ayed, E.~Cyffers, F.~Grimberg, C.~He, R.~Loeb,
  P.~Mangold, T.~Marchand, O.~Marfoq, E.~Mushtaq \emph{et~al.}, ``Flamby:
  Datasets and benchmarks for cross-silo federated learning in realistic
  healthcare settings,'' \emph{arXiv preprint arXiv:2210.04620}, 2022.

\bibitem{hsu2019measuring}
T.-M.~H. Hsu, H.~Qi, and M.~Brown, ``Measuring the effects of non-identical
  data distribution for federated visual classification,'' \emph{arXiv preprint
  arXiv:1909.06335}, 2019.

\bibitem{li2022federated}
Q.~Li, Y.~Diao, Q.~Chen, and B.~He, ``Federated learning on non-iid data silos:
  An experimental study,'' in \emph{2022 IEEE 38th International Conference on
  Data Engineering (ICDE)}.\hskip 1em plus 0.5em minus 0.4em\relax IEEE, 2022,
  pp. 965--978.

\bibitem{bland2004logrank}
J.~M. Bland and D.~G. Altman, ``The logrank test,'' \emph{Bmj}, vol. 328, no.
  7447, p. 1073, 2004.

\bibitem{archetti2023heterogeneous}
\BIBentryALTinterwordspacing
A.~Archetti, ``Federated survival datasets,'' 2023. [Online]. Available:
  \url{https://github.com/archettialberto/federated_survival_datasets}
\BIBentrySTDinterwordspacing

\bibitem{lee2003statistical}
E.~T. Lee and J.~Wang, \emph{Statistical methods for survival data
  analysis}.\hskip 1em plus 0.5em minus 0.4em\relax John Wiley \& Sons, 2003,
  vol. 476.

\bibitem{schumacher1994randomized}
M.~Schumacher, G.~Bastert, H.~Bojar, K.~H{\"u}bner, M.~Olschewski,
  W.~Sauerbrei, C.~Schmoor, C.~Beyerle, R.~Neumann, and H.~Rauschecker,
  ``Randomized 2 x 2 trial evaluating hormonal treatment and the duration of
  chemotherapy in node-positive breast cancer patients. german breast cancer
  study group.'' \emph{Journal of Clinical Oncology}, vol.~12, no.~10, pp.
  2086--2093, 1994.

\bibitem{dispenzieri2012use}
A.~Dispenzieri, J.~A. Katzmann, R.~A. Kyle, D.~R. Larson, T.~M. Therneau, C.~L.
  Colby, R.~J. Clark, G.~P. Mead, S.~Kumar, L.~J. Melton~III \emph{et~al.},
  ``Use of nonclonal serum immunoglobulin free light chains to predict overall
  survival in the general population,'' in \emph{Mayo Clinic Proceedings},
  vol.~87, no.~6.\hskip 1em plus 0.5em minus 0.4em\relax Elsevier, 2012, pp.
  517--523.

\bibitem{kaplan1958nonparametric}
E.~L. Kaplan and P.~Meier, ``Nonparametric estimation from incomplete
  observations,'' \emph{Journal of the American statistical association},
  vol.~53, no. 282, pp. 457--481, 1958.

\bibitem{nelson1972theory}
W.~Nelson, ``Theory and applications of hazard plotting for censored failure
  data,'' \emph{Technometrics}, vol.~14, no.~4, pp. 945--966, 1972.

\bibitem{aalen1978nonparametric}
O.~Aalen, ``Nonparametric inference for a family of counting processes,''
  \emph{The Annals of Statistics}, pp. 701--726, 1978.

\bibitem{cutler1958maximum}
S.~J. Cutler and F.~Ederer, ``Maximum utilization of the life table method in
  analyzing survival,'' \emph{Journal of chronic diseases}, vol.~8, no.~6, pp.
  699--712, 1958.

\bibitem{mcmahan2017communication}
B.~McMahan, E.~Moore, D.~Ramage, S.~Hampson, and B.~A. y~Arcas,
  ``Communication-efficient learning of deep networks from decentralized
  data,'' in \emph{Artificial intelligence and statistics}.\hskip 1em plus
  0.5em minus 0.4em\relax PMLR, 2017, pp. 1273--1282.

\bibitem{mothukuri2021survey}
V.~Mothukuri, R.~M. Parizi, S.~Pouriyeh, Y.~Huang, A.~Dehghantanha, and
  G.~Srivastava, ``A survey on security and privacy of federated learning,''
  \emph{Future Generation Computer Systems}, vol. 115, pp. 619--640, 2021.

\bibitem{rahimian2022practical}
S.~Rahimian, R.~Kerkouche, I.~Kurth, and M.~Fritz, ``Practical challenges in
  differentially-private federated survival analysis of medical data,'' in
  \emph{Conference on Health, Inference, and Learning}.\hskip 1em plus 0.5em
  minus 0.4em\relax PMLR, 2022, pp. 411--425.

\bibitem{li2020fedprox}
T.~Li, A.~K. Sahu, M.~Zaheer, M.~Sanjabi, A.~Talwalkar, and V.~Smith,
  ``Federated optimization in heterogeneous networks,'' \emph{Proceedings of
  Machine Learning and Systems}, vol.~2, pp. 429--450, 2020.

\bibitem{karimireddy2020scaffold}
S.~P. Karimireddy, S.~Kale, M.~Mohri, S.~Reddi, S.~Stich, and A.~T. Suresh,
  ``Scaffold: Stochastic controlled averaging for federated learning,'' in
  \emph{International Conference on Machine Learning}.\hskip 1em plus 0.5em
  minus 0.4em\relax PMLR, 2020, pp. 5132--5143.

\bibitem{acar2021federated}
D.~A.~E. Acar, Y.~Zhao, R.~M. Navarro, M.~Mattina, P.~N. Whatmough, and
  V.~Saligrama, ``Federated learning based on dynamic regularization,''
  \emph{arXiv preprint arXiv:2111.04263}, 2021.

\bibitem{cox1972regression}
\BIBentryALTinterwordspacing
D.~R. Cox, ``Regression models and life-tables,'' \emph{Journal of the Royal
  Statistical Society. Series B (Methodological)}, vol.~34, no.~2, pp.
  187--220, 1972. [Online]. Available:
  \url{http://www.jstor.org/stable/2985181}
\BIBentrySTDinterwordspacing

\bibitem{caldas2018leaf}
S.~Caldas, S.~M.~K. Duddu, P.~Wu, T.~Li, J.~Kone{\v{c}}n{\`y}, H.~B. McMahan,
  V.~Smith, and A.~Talwalkar, ``Leaf: A benchmark for federated settings,''
  \emph{arXiv preprint arXiv:1812.01097}, 2018.

\bibitem{lomurno2022sgde}
E.~Lomurno, A.~Archetti, L.~Cazzella, S.~Samele, L.~Di~Perna, and M.~Matteucci,
  ``{SGDE}: Secure generative data exchange for cross-silo federated
  learning,'' in \emph{AIPR 2022, International Conference on Artificial
  Intelligence and Pattern Recognition}, 2022.

\bibitem{kvamme2021continuous}
H.~Kvamme and {\O}.~Borgan, ``Continuous and discrete-time survival prediction
  with neural networks,'' \emph{Lifetime Data Analysis}, vol.~27, no.~4, pp.
  710--736, 2021.

\bibitem{katzman2018deepsurv}
J.~L. Katzman, U.~Shaham, A.~Cloninger, J.~Bates, T.~Jiang, and Y.~Kluger,
  ``Deepsurv: personalized treatment recommender system using a cox
  proportional hazards deep neural network,'' \emph{BMC medical research
  methodology}, vol.~18, no.~1, pp. 1--12, 2018.

\bibitem{ripley2023mass}
\BIBentryALTinterwordspacing
B.~Ripley, B.~Venables, D.~M. Bates, K.~Hornik, A.~Gebhardt, and D.~Firth, ``R
  package: Mass,'' Jul. 27, 2022. [Online]. Available:
  \url{https://stat.ethz.ch/R-manual/R-devel/library/MASS/html/00Index.html}
\BIBentrySTDinterwordspacing

\bibitem{therneau2023survival}
\BIBentryALTinterwordspacing
T.~Therneau, T.~Lumley, E.~Atkinson, and C.~Crowson, ``R package: survival,''
  Jan. 9, 2023. [Online]. Available:
  \url{https://stat.ethz.ch/R-manual/R-devel/library/survival/html/00Index.html}
\BIBentrySTDinterwordspacing

\bibitem{vanderbilt2022vanderbilt}
\BIBentryALTinterwordspacing
{Vanderbilt University Department of Biostatistics}, ``Vanderbilt biostatistics
  datasets,'' Dec. 1, 2022. [Online]. Available: \url{http://hbiostat.org/data}
\BIBentrySTDinterwordspacing

\end{thebibliography}
